\pdfoutput=1

\documentclass[11pt]{article}
\usepackage[table,dvipsnames]{xcolor}

\usepackage{EACL2023}
\usepackage{url}

\usepackage{times}
\usepackage{latexsym}
\usepackage[T1]{fontenc}

\usepackage[utf8]{inputenc}

\usepackage{microtype}

\usepackage{inconsolata}

\usepackage{graphicx}               
\usepackage{tabularx}               
\newcolumntype{C}{>{\centering\arraybackslash}X}
\usepackage{multirow}               
\usepackage{diagbox}                
\usepackage{hhline}                 
\usepackage{color}                  
\usepackage{amsmath,amsthm}                
\usepackage{bm}
\usepackage{amssymb}                
\usepackage{mathtools}              
\usepackage{enumitem}               
\usepackage{booktabs}
\usepackage{subcaption}
\usepackage{bbm}

\makeatletter
\newcommand*\bigcdot{\mathpalette\bigcdot@{.5}}
\newcommand*\bigcdot@[2]{\mathbin{\vcenter{\hbox{\scalebox{#2}{$\m@th#1\bullet$}}}}}
\makeatother

\usepackage{xparse}
\NewDocumentCommand{\heng}{ mO{} }{\textcolor{OrangeRed}{\textsuperscript{\textit{Heng}}\textsf{\textbf{\small[#1]}}}}
\NewDocumentCommand{\qianying}{ mO{} }{\textcolor{CadetBlue}{\textsuperscript{\textit{Qianying}}\textsf{\textbf{\small[#1]}}}}
\NewDocumentCommand{\fei}{ mO{} }{\textcolor{Blue}{\textsuperscript{\textit{Fei}}\textsf{\textbf{\small[#1]}}}}
\NewDocumentCommand{\lingfei}{ mO{} }{\textcolor{Cyan}{\textsuperscript{\textit{Lingfei}}\textsf{\textbf{\small[#1]}}}}
\NewDocumentCommand{\haoran}{ mO{} }{\textcolor{Brown}{\textsuperscript{\textit{Haoran}}\textsf{\textbf{\small[#1]}}}}

\NewDocumentCommand{\aysa}{ mO{} }{\textcolor{Purple}{\textsuperscript{\textit{Aysa}}\textsf{\textbf{\small[#1]}}}}

\usepackage{lipsum}  
\usepackage{longtable}
\usepackage{subcaption}
\usepackage{adjustbox}
\usepackage{tikz}
\usepackage{listings}

\colorlet{punct}{red!60!black}
\definecolor{background}{HTML}{EEEEEE}
\definecolor{delim}{RGB}{20,105,176}
\colorlet{numb}{magenta!60!black}

\newcommand{\mathnum}[1]{${#1}$}
\newcommand{\mathbnum}[1]{$\mathbf{#1}$}

\newcommand{\res}[2]{\begin{tabular}[t]{@{}c@{}}${#1}$\\ $^{\pm{#2}}$\end{tabular}}

\newcommand{\resb}[2]{\begin{tabular}[t]{@{}c@{}}$\mathbf{#1}$\\ $^{\pm{#2}}$\end{tabular}}

\lstdefinelanguage{json}{
    basicstyle=\scriptsize\ttfamily,
    numbers=left,
    numberstyle=\scriptsize,
    stepnumber=1,
    numbersep=8pt,
    showstringspaces=false,
    breaklines=true,
    frame=lines,
    backgroundcolor=\color{background},
    literate=
     *{0}{{{\color{numb}0}}}{1}
      {1}{{{\color{numb}1}}}{1}
      {2}{{{\color{numb}2}}}{1}
      {3}{{{\color{numb}3}}}{1}
      {4}{{{\color{numb}4}}}{1}
      {5}{{{\color{numb}5}}}{1}
      {6}{{{\color{numb}6}}}{1}
      {7}{{{\color{numb}7}}}{1}
      {8}{{{\color{numb}8}}}{1}
      {9}{{{\color{numb}9}}}{1}
      {:}{{{\color{punct}{:}}}}{1}
      {,}{{{\color{punct}{,}}}}{1}
      {\{}{{{\color{delim}{\{}}}}{1}
      {\}}{{{\color{delim}{\}}}}}{1}
      {[}{{{\color{delim}{[}}}}{1}
      {]}{{{\color{delim}{]}}}}{1},
}

\usepackage{xspace}
\usepackage{url}

\newcommand{\ours}{\textsc{ConEntail}\xspace}

\title{\ours: An Entailment-based Framework for Universal Zero and Few Shot Classification with Supervised Contrastive Pretraining}

\author{Ranran Haoran Zhang$^1$, Aysa Xuemo Fan$^2$, Rui Zhang$^1$ \\
$^1$ Penn State University \\
$^2$ University of Illinois at Urbana-Champaign\\
\texttt{\{hzz5361, rmz5227\}@psu.edu} \\
\texttt{xuemof2@illinois.edu}
}

\date{}

\begin{document}

\maketitle

\begin{abstract}
    A universal classification model aims to generalize to diverse classification tasks in both zero and few shot settings. 
    A promising way toward universal classification is to cast heterogeneous data formats into a dataset-agnostic ``meta-task'' (e.g., textual entailment, question answering) then pretrain a model on the combined meta dataset. 
    The existing work is either pretrained on specific subsets of classification tasks, or pretrained on both classification and generation data but the model could not fulfill its potential in universality and reliability. These also leave a massive amount of annotated data under-exploited.
    To fill these gaps, we propose \ours, a new framework for universal zero and few shot classification with supervised contrastive pretraining.
    Our unified meta-task for classification is based on nested entailment. It can be interpreted as ``Does sentence $a$ entails [sentence $b$ entails label $c$]''.
    This formulation enables us to make better use of 57 annotated classification datasets for supervised contrastive pretraining and universal evaluation. 
    In this way, \ours helps the model (1) absorb knowledge from different datasets, and (2) gain consistent performance gain with more pretraining data. 
    In experiments, we compare our model with discriminative and generative models pretrained on the same dataset. 
    The results confirm that our framework effectively exploits existing annotated data and outperforms baselines in both zero (9.4\% average improvement) and few shot settings (3.5\% average improvement). Our code is available 
    at \url{https://github.com/psunlpgroup/ConEntail}.

\end{abstract}  




\begin{figure*}[t]
  \centering
  \includegraphics[width=0.95\textwidth]{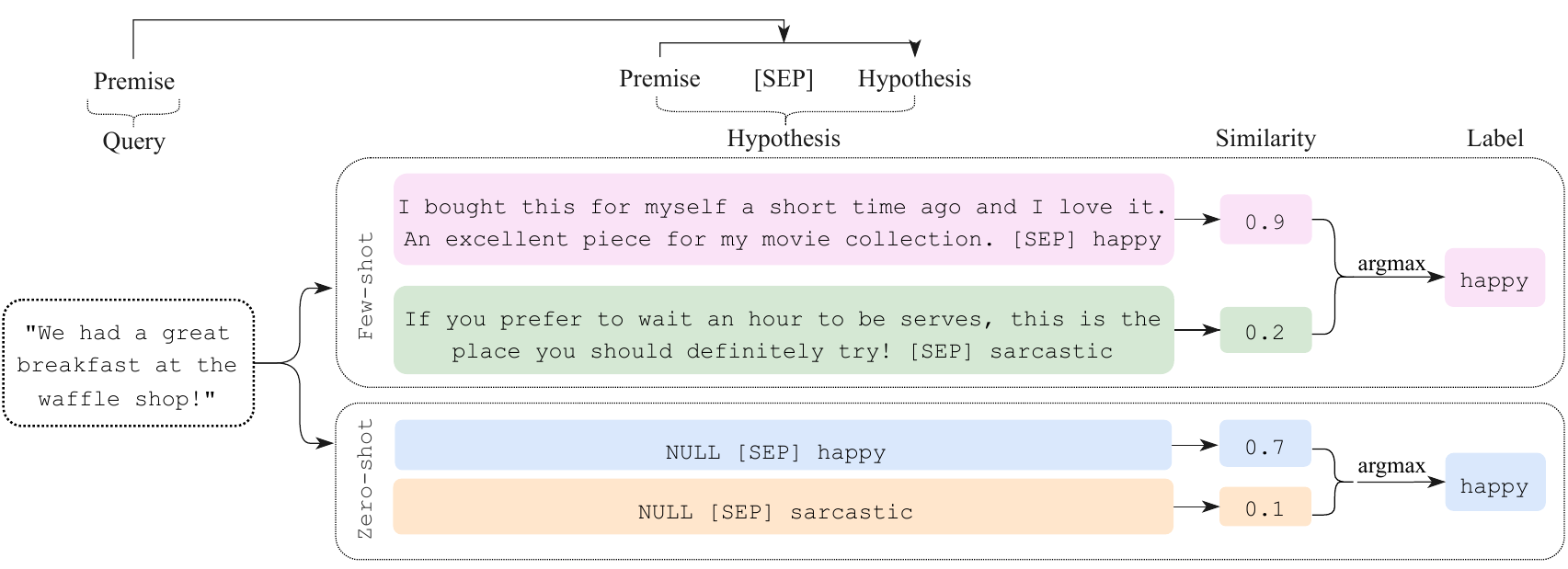}
  \caption{The overview of the \ours framework. By casting the classification as a nested entailment task, the model performs classification by telling if a query sentence $q$ entails [premise example $p$ entails hypothesis label $h$].
  In a few-shot setting, the premise is an example sentence; in a zero-shot setting, the premise is a ``NULL'' placeholder.
  }
  \label{fig:intro}
  \vspace{-1.5mm}
\end{figure*}

\section{Introduction}

It has been a long-standing effort to solve various text classification tasks by training one universal model~\cite{pmlr-v48-kumar16}.
With an ideal universal classification model, we can expect extreme generalization with few or zero annotation in new domains/tasks/datasets. 
To this end, researchers reformulate heterogeneous task definitions into a unified format of a meta-task in natural language \cite{yin2020universal,khashabi-etal-2020-unifiedqa}. 
Solving the meta-task is equivalent to solving the isolated tasks, thus the meta-task paves the way of supplementing unsupervised pretrained Language Models (PLM) with additional supervised pretraining, to further absorb knowledge from heterogeneous labeled data.


The success of universal classification models hinges on how well a strong PLM understands natural language meta-task.
The meta-task format depends on two underlying PLM types: 
(a) \textbf{discriminator} uses Encoder PLMs and treats all classification tasks as binary entailment classification problem \cite{yin2019benchmarking,yin2020universal,xia-etal-2021-incremental,wang2021entailment}.
However, they only pretrain models on Natural Language Inference datasets, whose knowledge is not comprehensive comparing all classification tasks \cite{ma2021issues}.
(b) \textbf{generator} uses Encoder-Decoder PLMs and treats all tasks as text generation problem \cite{gao2020making,raffel2020exploring,sanh2021multitask,aribandi2021ext5,ye2021crossfit,bragg2021flex,du2021all,schick-schutze-2021-exploiting,schick-schutze-2021-just}. 
Thus they are compatible with both classification tasks and generation tasks. 
However, the generator nature implies that the predicted texts may not match any possible labels, thus more likely to fail on classification tasks \cite{sanh2021multitask}.

Based on our observations and experiments, we argue that the discriminators have more potential in universal classification, and propose a new discriminator framework, \ours, that can make better use of existing annotated datasets.
Concretely, we reformulate the unified meta-task as a nested entailment: ``Does sentence $q$ entails [sentence $p$ entails label $h$]''.
Take Fig. \ref{fig:intro} as an example, the query ``We had a great breakfast at the waffle shop!'' entails the same label as the premise ``I bought this for myself a short time ago and I love it. An excellent piece for my movie collection.'', so it yields a high similarity score of 0.9, in this case, it is higher than any other similarities, thus, the prediction would be ``happy''.
For zero-shot generalization, as no annotated sentences are available, we replace the premise $p$ with ``NULL'' in evaluation. We randomly nullify a small ratio of $p$ in the supervised pretraining for training-evaluation consistency. 
The supervised contrastive learning framework pulls sentences embeddings with the same label together and pushes those with different labels apart, thus capturing more similarities/dissimilarities from labeled data, and benefiting few/zero-shot learning.

In experiments, we collect 56 classification datasets from Crossfit~\cite{ye2021crossfit}, together with their templates, to formulate a large supervised pretraining dataset. 
We reproduce EFL~\cite{wang2021entailment}, Unifew~\cite{bragg2021flex} and Crossfit~\cite{ye2021crossfit} in the same setting and control influences of PLM supervised pretraining data, then conduct fair comparison with our proposed \ours.  
The experiments show that generators (Unifew and Crossfit) do not fit the classification task well and thus significantly under-perform the random guess in zero-shot evaluation; standard discriminators (EFL) under-exploit supervised pretraining datasets and thus do not gain consistent improvement as pretraining data scale up, while \ours makes the best use of the supervised pretraining data and keep consistent performances. 
Our model outperforms baselines in both zero (9.4\% average improvement) and few shot settings (3.5\% average improvement). 

Our contributions are the following:
\begin{itemize}
    \item We propose a novel universal classification framework based on nested entailment, \ours, that can be used in both zero and few shot settings. It makes better use of supervised pretraining datasets and consistently improves performances with increases of the pretraining scale. 
    \item  We design systematic experiments to compare generative and discriminative models, and more importantly, we give in-depth analysis to reveal their attributes in universal classification task. 
    \item Our model reliably outperforms the baseline models in all kinds of pretraining size, fine-tuning size, and covers a wide range of tasks. 
\end{itemize}

\section{Related Work}


\noindent \textbf{Universal Meta Task}
Casting heterogeneous datasets into a unified meta-task allows researchers to train one model to solve all tasks.
There are two types of meta-task formats, 
generation~\cite{schick-schutze-2021-exploiting, schick-schutze-2021-just,gao2020making, ye2021crossfit, bragg2021flex, khashabi-etal-2020-unifiedqa} and discrimination~\cite{yin2019benchmarking,yin2020universal,xia-etal-2021-incremental,wang2021entailment}.
The generators formulate meta-task as a text-to-text generation problem. 
Although their supervised pretraining usually involves both classification and generation tasks, as the text outputs are open-ended, the model predictions may fall out of all possible labels. 
The discriminators formulate meta-task as an entailment classification problem, and usually use Natural Language Inference datasets for supervised pretraining. 
We extend discriminator pretraining to more classification datasets and propose a nested entailment meta-task to enable a more efficient supervised pretraining method. 

\noindent \textbf{Supervised Pretraining}
Supervised pretraining originates from explicit multitask learning~\cite{caruana1997multitask} which combines different task knowledge into shared representations. 
\citet{phang2018sentence} found that supplementing PLMs with supervised pretraining between unsupervised pretraining and downstream finetuning can significantly boost the performance and few-shot generalization. 
The discriminator models including UFO-Entail~\cite{yin2020universal} and EFL~\cite{wang2021entailment} 
are trained on MNLI~\cite{mnli} in a supervised fashion, but they do not combine different sources of datasets. 
Furthermore, T0~\cite{sanh2021multitask} and ExT5~\cite{aribandi2021ext5} extends T5~\cite{raffel2020exploring} by using 107 and 171 datasets for supervised pretraining and conduct zero-shot evaluation.
FLEX~\cite{bragg2021flex} and Crossfit~\cite{ye2021crossfit} extends the supervised pretraining evaluation to few-shot learning.

The supervised pretraining strategies from these works vary in pretraining datasets and hyperparameters, but they mostly follow their underlying language model tasks, such as Next Sentence Prediction or Text Generation. 
We argue that applying the unsupervised pretraining strategy to supervised pretraining is an underuse of the labeled data, and propose a supervised contrastive learning method on PLMs for better zero/few-shot generalization.

\noindent \textbf{Contrastive Learning for NLP}
Contrastive learning aims to create embeddings such that similar examples are close while dissimilar examples are far away \cite{1467314}. 
While most works use self-supervised contrastive learning~\cite{shen2020simple,fang2020cert,you-etal-2021-self-supervised,ye-etal-2021-efficient}, only a few adopt supervised contrastive learning.
CLIP~\cite{radford2021learning} uses labeled images and captions as supervision signal. SimCSE \cite{gao2021simcse} and SBERT \cite{reimers-2019-sentence-bert} use labeled sentence pairs from NLI to construct positive and negative examples. 
However, their contrastive data creations are limited to specific types of data, and thus can be hardly extended to universal classification. 
We reformulate all NLP classification tasks into a unified contrastive meta-task and use Supervised Contrastive Loss \cite{super_cl} to train on heterogeneous labeled data during supervised pretraining.





  \begin{figure}[t!]
  \centering
  \includegraphics[width=0.35\textwidth]{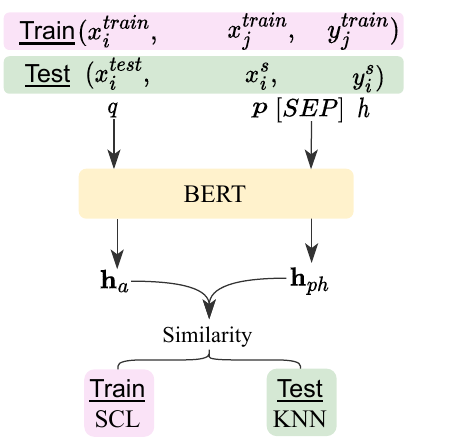}
  \caption{
  During supervised pertaining, the \ours model is optimized with pairwise contrastive learning loss SCL.
  Testing utilizes the K-Nearest Neighbor predictor to rank pairwise similarities between the query and premise-hypothesis pairs for retrieval of the most likely label.
  Zero-shot training/testing occurs when the premise example is represented by a "NULL" token."
  }
  \label{fig:overview}
  \vspace{-1.5mm}
  \end{figure}

\section{Method}
\subsection{Universal Classification}



Universal classification task aims to build a universal predictor that generalize to new domain/task/dataset based on only a few or zero newly annotated examples. 
In order for models to understand a new area, any available resources should be considered for learning, including PLMs trained on large-scale unsupervised data and heterogeneous supervised classification datasets in the NLP community. 
To leverage heterogeneous datasets, the disparate input-output formats need to be reformulated to a unified PLM comprehensible format, i.e., ``meta task'', through either human-curated or machine-generated templates.
Then a universal model on the combined meta dataset is trained, which applies universal predictors to new areas. 
Because the meta task format is compatible with every task, we can cast target tasks into the same format, in this way solving the meta task is equivalent to solving tasks in a new area.

\subsection{\ours: Nested Entailment}

In this paper, we introduce a supervised contrastive pretraining paradigm that makes better use of supervised pretraining. The overview is shown in Fig. \ref{fig:overview}.
Our \ours model takes 3 inputs: 
\begin{align*}
f \colon \mathcal{Q}, \mathcal{P},\mathcal{H}  &\to  \{0, 1\} \\
    q, p, h &\mapsto b
\end{align*}
where $q \in \mathcal{Q}$ is the \textbf{query} sentence to be classified. $p \in \mathcal{P}$ is the exemplar sentence as a \textbf{premise}, $h  \in \mathcal{H}$ is the \textbf{hypothesis} verbalized from the label of $p$. 
The task of \ours is to determine \textbf{if $q$ entails [$p$  entails $h$]}.

We follow  \cite{khashabi2020unifiedqa,ye2021crossfit} and translate sentence and label $(x, y)$ to $(q, p, h)$ in a PLM comprehensible format, e.g.,
\begin{itemize}[noitemsep]
    \item $x \mapsto q$, where $q$ is the input sentence $x$ with multiple-choice, for example, \textit{(1) happy (2) sarcastic (3) sad, sentence: I bought this  for myself ...}
    \item $x \mapsto p$: where $p$ is the input sentence $x$ with premise, for example, \textit{sentence: I bought this for myself ...}
    \item $y \mapsto h$ where $h$ is the label name, for example, $h$: \textit{happy}
\end{itemize}
where we provide $q$ with all possible labels as multiple-choice questions, and concatenate them in a linearized sentence. 
In supervised pretraining, $q$ and $p$ are two different surface forms of the same $x$, so that we can construct positive and negative examples for the later contrastive learning. In the test, $q$ is the query sentence to be clarified and $p$ and $h$ are from the support set.




We use $\text{BERT}_{\text{base}}$ to encode sentences to vector representation $\mathbf{h}$. 
\begin{equation}
    \mathbf{h}_{q} = \text{BERT}_{\text{base}}(q) \\
\end{equation}

$p$ and $h$ are then concatenated into one sequence to be fed into the encoder:
\begin{align}
    ph &= p \texttt{[SEP]} h \\
    \mathbf{h}_{ph} &= \text{BERT}_{\text{base}}(ph)
\end{align}


In the supervised pretraining,
the embeddings of each mini-batch are  composed by $\left\{\mathbf{h}_{q}^i, \mathbf{h}_{ph}^i\right\}_{i=1,...,N}$, where  $N$ is the batch size. 
Then we calculate their pairwise cosine similarity $\text{sim}\left({\mathbf{h}_{q}^i, \mathbf{h}_{ph}^j}\right) = \frac{\mathbf{h}_{q}^i \cdot \mathbf{h}_{ph}^j}{\|\mathbf{h}_{q}^i \| \cdot \|\mathbf{h}_{ph}^j \|} $ for contrastive training.
$s_{ij} \in \left\{0, 1\right\}$ is denoted as the groundtruth of the predicted similarity, where $s_{ij} = 1$ is a positive pair when $y_i = y_j$,
and vice versa. 
The positive/negative examples are constructed by all combinations of instances in the batch, note that we did not mine hard examples. 
We follow the balanced sampling strategy from Meta Classification Learning \cite{mcl19} that each label in a mini-batch has an equal number of input sentences. 
In the test phase, we calculate cosine similarities between $q$ and all possible $ph$ and output the most similar $h$ as the prediction result. 
Thus, we consider our setting as a K-way N-shot learning, where K is determined by the test set, N varies from 0 to 80 in our experiments.

Given the pairwise similarity, we use Supervised Contrastive Loss \cite{super_cl} to train the model:
\begin{equation}
\begin{split}
  \mathcal{L}
  =-&\sum_{i=1}^{N}\frac{1}{|P(i)|}\sum_{p=1}^{N}\mathbbm{1}_{y_i=y_p}\mathbbm{1}_{i\neq p} \\
  &\log{\frac{\exp\left(\text{sim}\left({\mathbf{h}_{q}^i, \mathbf{h}_{ph}^p}\right)/\tau\right)}{\textstyle\sum_{a=1}^{N}\mathbbm{1}_{i\neq a}\exp\left(\text{sim}\left({\mathbf{h}_{q}^i, \mathbf{h}_{ph}^a}\right)/\tau\right)}}
\end{split}
  \label{eq:supervised_loss}
\end{equation}
where 
$|P(i)| = \sum_{p=1}^N \mathbbm{1}_{y_p = y_i}$ is the number of all positive pairs, $\tau$ is the temperature hyperparameters.
Different from self-supervised contrastive learning losses, such as SimCSE \cite{gao2021simcse}, the positive pairs in Supervised Contrastive Loss can be more than one.

To enable zero-shot generalization, inspired by BERT masked language model \cite{devlin-etal-2019-bert}, we introduce a dummy premise ``NULL'' in both supervised pretraining and testing. 
During supervised pretraining, we randomly replace 5\% of the premise $p$ with ``NULL'' (\textbf{if $q$ entails [``NULL'' entails $h$]}.).
During zero-shot test, the support set is empty and the model uses only ``NULL'' and label names to answer the question.

\begin{table*}[t!]
\resizebox{\textwidth}{!}{
\normalsize
\centering
\begin{tabular}{lcccc|ccccc|c}
\toprule
 \textbf{Method} & \textbf{CoLA} & \textbf{QQP} & \textbf{Hate\_speech} & \textbf{MRPC} & \textbf{SCITAIL}&  \textbf{Amazon} & \textbf{AGNews} & \textbf{Rotten\_tomatoes} & \textbf{SST-2} & \textbf{AVG} \\ 
                      & \multicolumn{4}{c|}{\textit{Unseen}}                                                                                                                                                                                    & \multicolumn{5}{c}{\textit{Seen}}                                                                                                                                                                                                                                                                  & \multicolumn{1}{|c}{}                                \\

\midrule
$\text{Random-guess}$  & \mathnum{50.5} & \mathnum{49.8} & \mathnum{34.1} &  \mathnum{50.0} & \mathnum{49.8} & \mathnum{49.9} & \mathnum{24.0} & \mathnum{46.8} & \mathnum{49.9} & \mathnum{44.9} \\

\midrule
\multicolumn{2}{l}{\textit{$0$-shot}}\\
\midrule
$\text{Crossfit}^\dagger$        &       \mathnum{0.0}   &       \mathnum{0.0}   &       \mathnum{0.0}  &       \mathnum{0.0}   &       \mathnum{0.2^*}   &       \mathnum{9.9^*}   &       \mathnum{0.0}   &       \mathnum{59.9^*}  &       \mathnum{33.4^*}   &       \mathnum{11.5^*}  \\
$\text{Unifew}^\dagger$  &       \mathnum{0.0}   &       \mathnum{0.0}   &       \mathnum{0.0}  &       \mathnum{0.0}   &       \mathnum{48.4^*}  &       \mathnum{63.7^*}  &       \mathnum{8.0^*}   &       \mathnum{57.4^*}  &       \mathnum{60.6^*}   &       \mathnum{26.5^*}  \\
$\text{EFL}$        & \mathbnum{62.6}  &       \mathbnum{60.5^*}  &       \mathnum{12.7}   & \mathnum{33.1}  &       \mathnum{47.2^*}  &       \mathnum{71.9^*}  &       \mathbnum{60.8^*}  &       \mathnum{72.5^*}  &       \mathnum{79.1^*}  &       \mathnum{53.8^*}  \\
$\text{\ours}$ &       \mathnum{58.5^*}  &       \mathnum{45.3}  &       \mathbnum{78.3^*}  &       \mathbnum{58.1^*}  &       \mathbnum{68.7^*}  &       \mathbnum{89.7^*}  &       \mathnum{52.8^*}  &       \mathbnum{78.1^*}  &       \mathbnum{83.0^*}   &       \mathbnum{63.2^*}  \\
\midrule
\multicolumn{2}{l}{\textit{$10$-shot fine-tuning}}\\

\midrule
$\text{Crossfit}^\dagger$        &       \res{55.3}{5.0}         &       \res{53.4}{9.8}         &       \res{42.8}{14.4}        &       \res{60.0}{11.1}        &       \res{58.8}{5.4}         &       \res{87.9}{6.1}         &       \res{83.7}{6.6}         &       \res{75.8^*}{1.2}        &       \res{81.2}{8.9}        &   \mathnum{65.3}  \\
$\text{Unifew}^\dagger$  &       \res{49.0}{4.9}         &       \resb{60.4}{6.0}         &        \res{34.9}{6.8}        &       \res{57.7}{6.3}         &       \res{53.4}{2.4}         &       \res{88.8}{3.6}         &       \resb{86.5}{1.8}         &       \res{73.4}{9.5}         &      \res{71.2}{11.5}         &       \mathnum{63.9}  \\
$\text{EFL}$       &       \resb{63.7}{0.2}         &       \resb{60.4}{0.2}         &       \res{13.8}{0.6}         &       \res{33.1^*}{0.0}         &       \res{47.2^*}{0.1}         &       \res{72.0}{0.0}         &       \res{62.3}{0.6}         &       \res{72.5^*}{0.0}         &       \res{79.5}{0.2}          &   \mathnum{55.9}  \\
$\text{\ours}$  &       \res{60.5^*}{0.6}        &       \res{55.6}{3.5}        &       \resb{44.7}{2.2}       &       \resb{69.9^*}{0.9}        &       \resb{71.0^*}{0.9}        &       \resb{89.4^*}{0.1}      &       \res{70.3^*}{2.1}        &       \resb{78.7^*}{0.2}        &       \resb{83.2^*}{0.2}        &       \mathbnum{68.8^*}  \\
\bottomrule
\end{tabular}
}
\caption{The main results of \ours compared with baselines. 
$\dagger$ indicates the models are generative models and the others are discriminative models.  
In the 10-shot evaluation, to offset the high variances from fine-tuning on such a small support set, the models are fine-tuned by 3 different random sampled support sets. After conducting experiments with and without supervised pretraining, we report the mean accuracy scores and the standard deviation of the best versions of models (in \textbf{bold}). We split the test sets in two groups, \textit{seen} and \textit{unseen}, which indicates if the test label names have occurred in the supervised pretraining. AVG is the highest average score of the two versions of models. If a model with supervised pretraining is better than that without supervised pretraining, it is indicated with a $*$. 
}
\label{tab:main}
\vspace{-1.5mm}
\end{table*}

\section{Experiments}
In this section, we describe our experiment setups including dataset selection, evaluation, and baseline models.

\subsection{Dataset Selection}
For universal text classification, we aim to cover the most popular text classification tasks, such as topic classification, sentiment analysis, paraphrase identification, and natural language inference.
Therefore, we adopt Crossfit~\cite{ye2021crossfit} that provides abundant hand-craft templates covering 56 classification tasks as the source of supervised pretraining and testing. 
We select 47 datasets as supervised pretraining sets and 9 widely accepted datasets as test sets: CoLA~\cite{cola}, QQP~\cite{qqp}, SST-2~\cite{sst2}, MRPC~\cite{mrpc}, SCITAIL~\cite{scitail}, Amazon Polarity~\cite{amazon}, AGNews~\cite{zhang2015character}, Rotten\_tomatoes~\cite{rotten}, Hate\_speech\_offensive~\cite{hateoffensive}. 
For the sentence-pair datasets (e.g., QQP, SST-2, MRPC), we adopt the Crossfit method by concatenating the two sentences with \texttt{[SEP]} to form one sequence for either $q$ or $p$.
From the 47 datasets for supervised pretraining, we randomly select 128 annotated examples per label.
As the same label name may occur in different datasets, to investigate the effect of label name overlapping, we pick 5 (out of 9) selected test sets with overlapping/seen label names for the supervised pretraining. 
The detailed dataset list is in Appendix~\ref{sec:appendix}. 

\subsection{Evaluation}


\noindent \textbf{Supervised Pretraining} To investigate the effect of the supervised pretraining, we consider two versions of all the compared models: (1) without supervised pretraining: we apply the original PLMs directly to the reformulated input-output test set. (2) with supervised pretraining: we first perform supervised pretraining on the PLMs and then evaluate the models with the updated parameters. 

\noindent \textbf{Zero-shot Evaluation} 
In zero-shot evaluation, the only available resources for the target task are the possible label names and the whole test set
will be used to evaluate the model.

\noindent \textbf{Few-shot Evaluation}
In few-shot evaluation, in addition to the label names, a small support set are available for fine-tuning the universal classification model.
The support set for each dataset is composed by $k$ random sampled annotated examples per label, from the training data. 
With small support sets, the evaluation score may have huge variance, thus we fine-tune and evaluate the model with 3 different support sets and report the mean and standard deviation.


\subsection{Baseline Models}


We aim to evaluate models in different paradigms in the same universal classification experiment setting. 
To this end, we compare three baselines that are most representative of the current literature on generators and discriminators. 

In this paper, we only consider the differences of the baselines in the meta-task formulation and their generator/discriminator nature while keeping other factors the same, so 
we reproduce the baselines strictly follow this rule, and use a similar size of pretrained language models as backbones, for a fair comparison. 
Because our generator/discriminator taxonomy suits many other existing works, with only subtle differences either in the templates or in the backbone PLMs from the baselines mentioned here, we do not add more baselines for comparisons. 


\noindent \textbf{Crossfit}~\cite{ye2021crossfit}: A generative model uses an  encoder-decoder structure. The encoder takes the query sentence, and the decoder generates the label name.

\noindent \textbf{Unifew}~\cite{bragg2021flex}: A generative model concatenates all possible labels to the input sentence as multiple-choice question answering. It uses an encoder-decoder structure and generates the label names as answers. 

\noindent \textbf{EFL}~\cite{wang2021entailment}: A discriminative model reformulates the tasks as multiple entailment binary classifications. Both the query sentence and the label name are fed into the encoder. The embedding of \texttt{[CLS]} token is used for binary classification. The label with the highest probability is the predicted output. 
For supervised pretraining, we enumerate all possible labels for input and provide all the ground truths for the binary classification.

\begin{figure}[t!]
  \centering
  \includegraphics[width=0.49\textwidth]{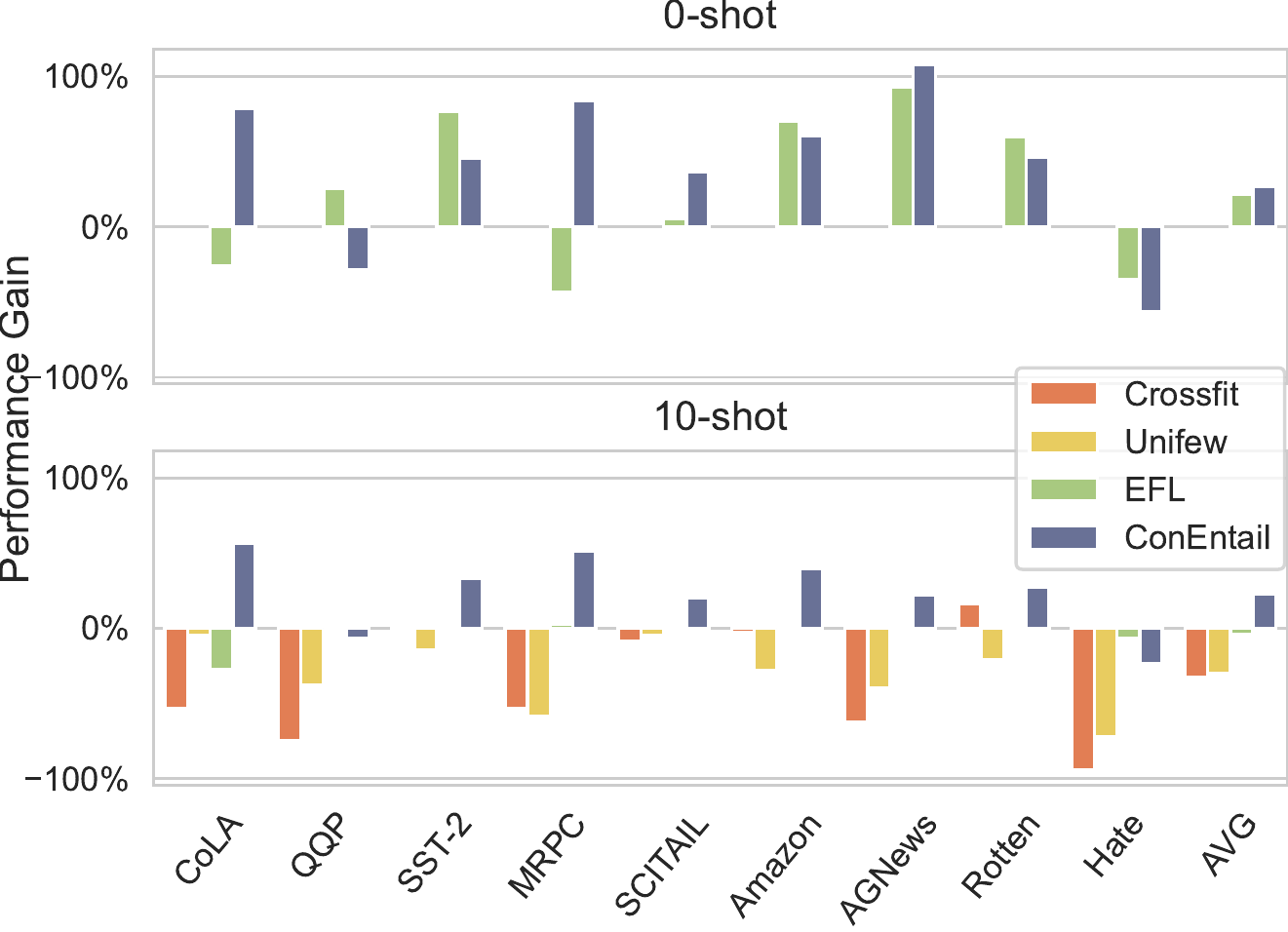}
  \caption{Relative performance gain of supervised pretraining on different datasets and models. The setting is the same with the main experiment. We do not plot zero-shot gains for the generators because most scores are 0 before and after supervised pretraining.}
  \label{fig:supervised_barplot}
  \vspace{-1.5mm}
\end{figure}

\begin{figure}[t!]
  \centering
  \includegraphics[width=0.48\textwidth]{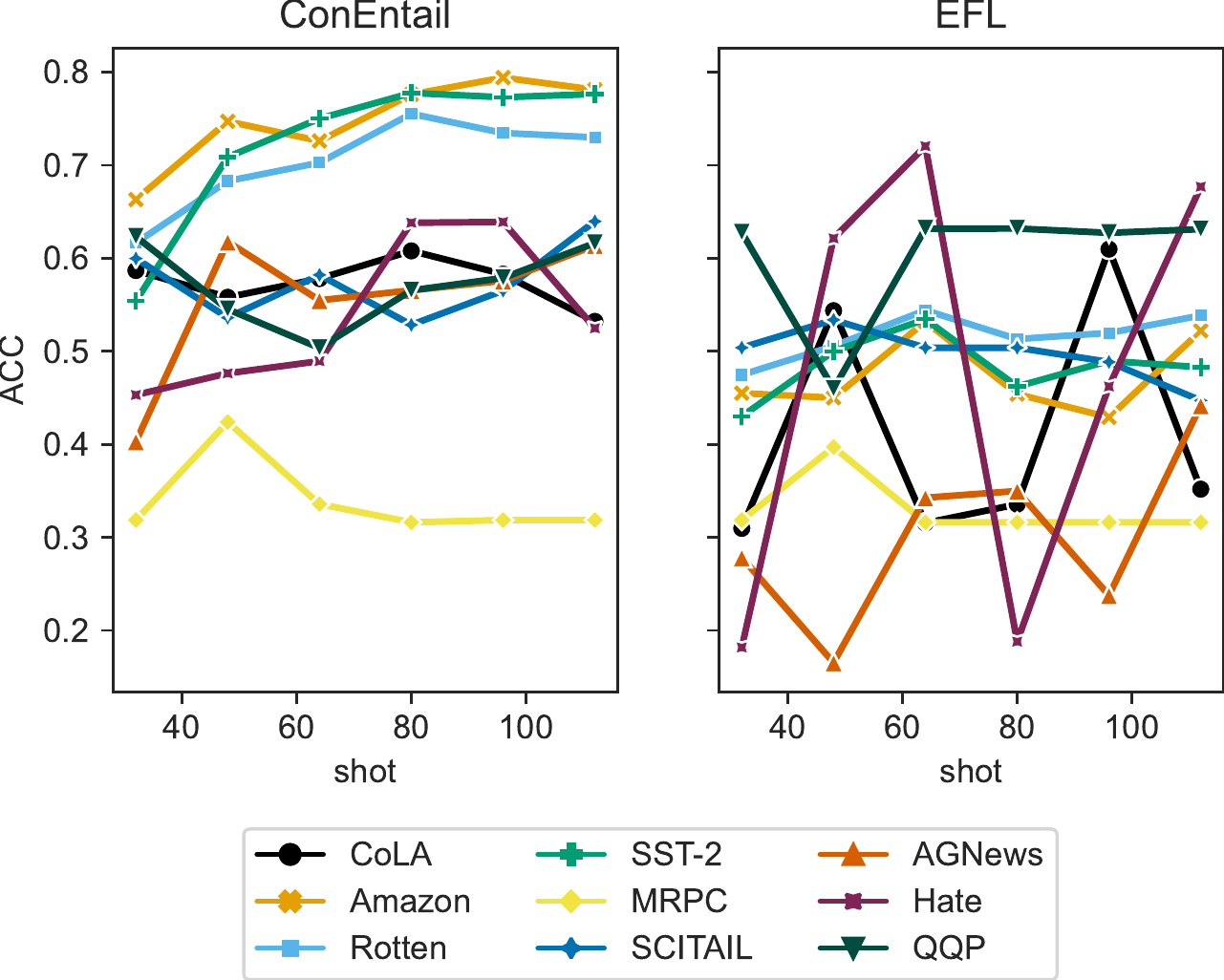}
  \caption{The effect of supervised pretraining data size. We show the zero-shot performance of \ours and EFL using different pretraining data size from 32 to 128 annotated sentences per label.
  }
  \label{fig:train32-128}
  \vspace{-1.5mm}
\end{figure}

\section{Results and Analysis}     
We design the following experiments to demonstrate and analyze the effectiveness of our method.
First, we present the best scores of the compared models with or without supervised pretraining as our main result (Section~\ref{sec:main}). 
Then, we investigate the performance gain or loss of each model brought by the supervised pretraining (Section~\ref{sec:supervised_pretraining}). 
Furthermore, we study the fine-grained impact of more labeled data in supervised pretraining or of more labeled data in support set (Section~\ref{sec:impact_more_training}). 
Considering these results, we discuss the difference between discriminators and generators (Section~\ref{sec:diff}).
Finally, we show a case study of universal classification under a zero-shot scenario (Section~\ref{sec:case}).

\subsection{Main Results}
\label{sec:main}


\begin{figure*}[t!]
  \centering
  \includegraphics[width=0.96\textwidth]{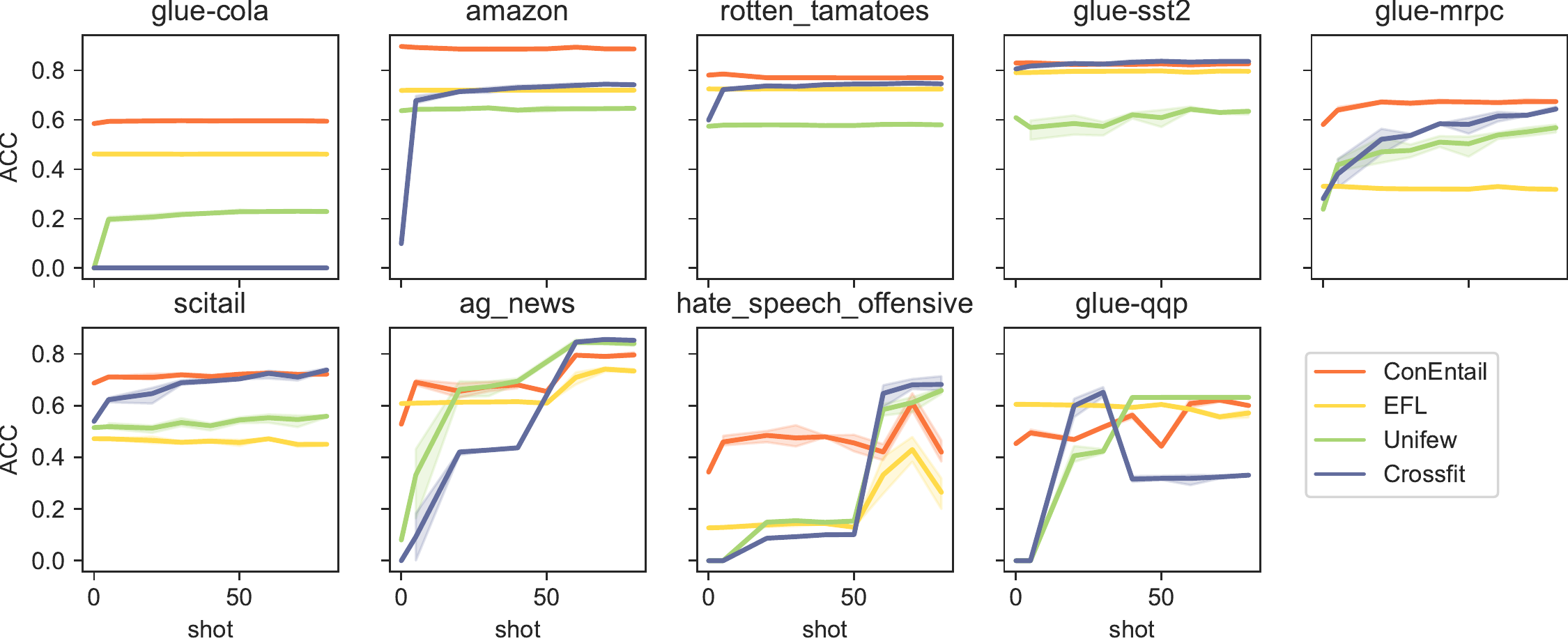}
  \caption{The effect of data size in the support set. We show the accuracy of the compared models fine-tuned with 0 to 80 examples in the support set. For each data size, we randomly sample 3 support sets for fine-tuning and evaluate on the same test set. 
  }
  \label{fig:finetune0-100}
  \vspace{-1.5mm}
\end{figure*}

We evaluate the models in two scenarios, 0-shot learning and 10-shot learning (Table \ref{tab:main}).
The average performances of both discriminator models, EFL and \ours, significantly outperform random guess and two generation-based models.
Particularly, \ours, with significantly improved average results, performs the best on 6 out of the 9 datasets in both 0-shot and 10-shot settings.

From the table, we also observe that the seen labels bring most improvements to Unifew in 0-shot setting. The 0-shot performance of Unifew in SST-2, SCITAIL and Amazon is far better than Crossfit. This is because Unifew has included the labels in the query sentences as multiple-choice questions, which provides the model additional familiarities from the supervised pretraining.
In other words, although the 0-shot unseen accuracies of the generative models are mostly 0,
their performances can be improved quickly with few-shot finetuning.
This indicates that generative models are promising few-shot learners but not strong zero-shot learners.

\subsection{Performance Gain from Supervised Pretraining}
\label{sec:supervised_pretraining}
We then quantify the effect of supervised pretraining by Relative Performance Gain introduced~\cite{ye2021crossfit}. Relative Performance Gain is the relative improvement brought by the supervised pretraining. It is defined as $\frac{\text{Acc}_\text{w} - \text{Acc}_{\text{w/o}}}{\text{Acc}_\text{{w/o}}}$, the performance difference between a supervised pretraining model $\text{Acc}_\text{w}$ and non-supervised pretraining model $\text{Acc}_\text{{w/o}}$, divided by the latter.
The results are shown in Fig.~\ref{fig:supervised_barplot}.

We observe that supervised pretraining boosts the performance in most datasets in the 0-shot setting. But it lowers the scores in the 10-shot setting, except for \ours. \ours's performance rises in 7 out of 9 datasets in both 0-shot and 10-shot setting. 
This shows the general necessity of supervised pretraining for 0-shot evaluation and the effectiveness of our proposed model in both settings.
The baseline models did not benefit from supervised retraining for the 10-shot setting because their conventional fine-tuning strategy is less likely to thoroughly update the parameters than our proposed contrastive learning. Noting that 10-shot evaluation means all the compared models only have 10 labeled examples for finetuning. 
\subsection{Impact of More Training data}
\label{sec:impact_more_training}
\noindent \textbf{More data in supervised pretraining}:
we investigate if more labeled data in supervised pretraining can improve zero-shot generalization. 
As the accuracies of generator models are close to zero in the zero-shot setting, we only consider discriminator models including \ours and EFL.
These two models are supervised pretrained on different-scale datasets (32-128 sentences per label) and evaluated on the 9 test sets.
As shown in Fig.~\ref{fig:train32-128}, the performance of \ours has fewer fluctuations than the EFL, and the performance improvements of most datasets flat after 80 shots for \ours. 
This observation implies that the supervised pretraining has significant and reliable positive effects on \ours with merely a small amount of supervised dataset.

\noindent \textbf{More data in support set}:
for models supervised pretrained with 128 annotated sentences per label, we plot the line chart of fine-tuning with 0 to 80 shots. 
As shown in Fig.~\ref{fig:finetune0-100}, adding a few training sentences may not largely boost performance when the universal model is strong enough, but it improves the models significantly if the models have a slow start.
Furthermore, though the generator model performances improve fast from 0 to 50 shots, the scores fluctuate largely. But after the first 50 shots, the improvements slow down, and the variances becomes much smaller. 
This implies that all the compared models are strong few shot learners, so that fine-tuning on large-scaled training data in the downstream tasks is unnecessary.


\subsection{Discussion on the Differences Between Discriminator and Generator Models}
\label{sec:diff}

\begin{table*}[h]
\centering
\resizebox{0.84\textwidth}{!}{
\small         
\extrarowheight=\aboverulesep     \addtolength{\extrarowheight}{\belowrulesep}     \aboverulesep=1pt     \belowrulesep=1pt \begin{tabular}{p{0.090000\textwidth} p{0.030000\textwidth} p{0.090000\textwidth} p{0.030000\textwidth} p{0.090000\textwidth} p{0.030000\textwidth} p{0.090000\textwidth} p{0.030000\textwidth} p{0.090000\textwidth} p{0.030000\textwidth}} \toprule

\multicolumn{10}{p{0.9\textwidth}}{I happily donate any covid vaccine dose which may be reserved for me to any person that is stupid enough to get one, or two, three, or four.} \\
\cellcolor{gray!28.946747}mild & \cellcolor{gray!28.946747}0.59 & \cellcolor{gray!17.930467}irony & \cellcolor{gray!17.930467}0.48 & \cellcolor{gray!13.626842}happy & \cellcolor{gray!13.626842}0.44 
& ... & ... &optimism & 0.23 \\

\midrule
\multicolumn{10}{p{0.9\textwidth}}{Guys it's OK. Delta says covid is over. IT'S OK NOW.} \\
\cellcolor{gray!39.272637}mild & \cellcolor{gray!39.272637}0.69 & \cellcolor{gray!18.454332}non-irony & \cellcolor{gray!18.454332}0.48 & \cellcolor{gray!16.634850}irony & \cellcolor{gray!16.634850}0.47 &
... & ... &hate & 0.10 \\

\midrule
\multicolumn{10}{p{0.9\textwidth}}{The first patient who died of COVID in Kerala already had BP and cardiac issues,  and he was 69.
Bottomline : If we take precautions,  we are still safe and can ensure others are safe too. } \\
\cellcolor{gray!21.274484}optimism & \cellcolor{gray!21.274484}0.51 & \cellcolor{gray!21.112837}mild & \cellcolor{gray!21.112837}0.51 & \cellcolor{gray!12.624205}positive & \cellcolor{gray!12.624205}0.43 &
... & ... &hate & 0.08 \\
\midrule

\multicolumn{10}{p{0.9\textwidth}}{Could you imagine putting your faith into the narrative, getting jabs, getting sick from side effects (which is now being called the variant) and then being labeled an antivaxxer amidst this lie "only the unnvacinated are getting sick". They will use you up until there's nothing left.} \\
\cellcolor{gray!28.631748}offensive & \cellcolor{gray!28.631748}0.59 & \cellcolor{gray!28.027834}irony & \cellcolor{gray!28.027834}0.58 & \cellcolor{gray!25.585313}mild & \cellcolor{gray!25.585313}0.56 &
... & ... &happy & 0.25 \\


\midrule
\multicolumn{10}{p{0.9\textwidth}}{...
I don't see a monetary benefit.
I don't see any professional benefit.
Ask the people who believe what they are being told for an explanation because I don't see any.} \\
\cellcolor{gray!29.892756}offensive & \cellcolor{gray!29.892756}0.60 & \cellcolor{gray!23.764319}mild & \cellcolor{gray!23.764319}0.54 & \cellcolor{gray!19.251336}irony & \cellcolor{gray!19.251336}0.49 &
... & ... &optimism & 0.20 \\

\midrule

\multicolumn{10}{p{0.9\textwidth}}{I can't do this anymore. I went from a house and 2 beautiful daughters and wife to homeless and left with literally nothing.
...
 They need to die painfully and even then they will never pay for their sins.
All it takes it one moment in history for everything to change. You keep breaking men down to nothing.
Those broken men will break you.} \\
\cellcolor{gray!49.852635}offensive & \cellcolor{gray!49.852635}0.80 & \cellcolor{gray!33.289273}negative & \cellcolor{gray!33.289273}0.63 & \cellcolor{gray!29.123498}hate & \cellcolor{gray!29.123498}0.59 &
... & ... &optimism & 0.10 \\


\bottomrule
\end{tabular}}
\caption{Case study of an unseen task. We use \ours in a zero-shot manner to analyze twitter and reddit sentiment during the Covid-Omicron surge. We pick 13 fine-grained sentiment labels and rank the labels by their similarity with the input sentence.}
\label{tab:case}
\vspace{-2mm}
\end{table*}

The ineffectiveness of zero-shot Unifew and Crossfit are rooted in their generation nature. 
The original motivation of generation-based models is to resolve all kinds of NLP tasks, including both classification and generation. 
However, the universal classification task (i.e., tasks in this paper) are usually formulated as label picking from limited choices, while generation tasks aim to output human-readable sentences that match the input sentences -- the target distributions for these 2 tasks are innately different. 
In the few-shot setting, finetuning with 10 more examples in the target task shifts the text generation distribution towards the label distribution, so the generated texts are more likely to be the labels, and this improves model performances.
However, as the predictions are still in the large vocabulary space, they are likely to be altered by any disturbances.   
When using different support sets, the variances of the accuracy are far larger than that of the discriminator models.
This also explains why Unifew performs better than Crossfit: the only difference between Unifew and Crossfit is that the input sentences of Unifew are appended with all possible label texts. By providing the generation process label hints, Unifew shifts its generation distribution towards label distribution and outperforms Crossfit.
But the accuracy gap between Unifew and Crossfit drops from 15\% to merely 0.7\% while the number of shots increases from 0 to 10. As we stated before, Unifew performs better in the 0-shot setting because of its extra label hints. However, with an increase of shots, this advantage is diluted, resulting in a smaller performance difference between these two models.

\subsection{A Case Study of Universal Classification}
\label{sec:case}
Consider a possible application scenario of universal classification:
when dealing with new tasks and domains, especially related to newly emerged events, usually people only have the label names in hand. 
Based on this, we demonstrate a COVID-19 sentiment classification case study to show the universality of the proposed \ours model.

We use keywords to collect 50 sentences from Reddit and Twitter during the surge of the Omicron variant, then pick 13 fine-grained sentiment labels for this task: \textit{positive, mild, negative, offensive, happy, anger, sad, hate, irony, non-offensive, non-irony, non-hate, optimism}. 
For each COVID-related query sentence, \ours model retrieves from all 13 possible labels and ranks them by similarity. 

From the results Table~\ref{tab:case} we observe that the model ranks the labels correctly most of the time.
With antonyms paired with each other, such as \textit{hate}/\textit{non-hate} and \textit{happy}/\textit{sad}, our model successfully predicts the labels with only the label names, showing the polarity derived from the pairwise ranking are effective and reliable.

\section{Conclusions}

In this work, we study the universal classification problem, that leverages heterogeneous labeled datasets to benefit zero/few-shot learning in a new domain/task/dataset.  
We conduct systematic experiments on mainstream discriminators and generators models, thoroughly evaluate different models, reveal their innate properties of meta-task reformulation and supervised pretraining strategies.
The results show that the generators with open-end prediction fail in zero-shot learning and the discriminators with a standard entailment meta-task hardly obtain a performance boost when more pretraining data is available. 
Our work provides a new angle for future researchers to explore universal NLP,
and propose a new nested entailment meta-task and a supervised contrastive learning strategy, \ours, to make better use of widely available annotated datasets, and adapts to new datasets with limited resources.  

\newpage
\section*{Limitations}

Although this paper aims to improve the universal generalization in the classification task, there are several limitations: 
(1) We do not compare with cloze-based models \cite{schick-schutze-2021-exploiting,schick-schutze-2021-just,gao2020making}, because their templates are more complicated and hard to be reproduced with our current datasets. 
(2) We do not consider structural classification tasks, such as Named Entity Recognition and Relation Extraction. 
(3) We only take classification datasets into account because our implementation is restricted by huggingface datasets and human-curated templates. We plan to extend our framework to more datasets in the future. 
(4) Due to the constraints from the templates and datasets, the class number of each test set is below 10. We plan to extend our framework to more labels in the future work.
(5) The compatibility of knowledge in similar tasks is assumed, but this assumption may not hold true due to varying annotation standards across datasets. For instance, MRPC and QQP are both paraphrase identification tasks, but MRPC uses hard example mining techniques, resulting in longer and more sophisticated sentences than QQP.
(6) The current study is limited to English datasets and can be extended to multiple languages in the future by using multilingual PLMs and pretraining datasets.

\section*{Acknowledgments}
We thank Qianying Liu for her valuable discussion. 

\bibliography{ref}
\bibliographystyle{acl_natbib}

\clearpage

\appendix
\onecolumn

\section{Hyperparameters and Implementation Details}
Unifew and Crossfit, as generative models, use $\text{BART}_{\text{base}}$ \cite{lewis-etal-2020-bart} as the backbone language model. In the supervised pretraining, we use AdamW optimizer \cite{loshchilov2017decoupled} with learning rate 3e-5, warm-up ratio 0.6\% and linear decay. In the meta-testing, we use the same hyperparameters and train 400 epochs for finetuning.

EFL and Entail2, as discriminator models, use $\text{BERT}_{\text{base}}$ \cite{devlin-etal-2019-bert} as the backbone language model.  In the supervised pretraining, we use AdamW optimizer \cite{loshchilov2017decoupled} with learning rate 1e-5, warm-up ratio 6\% and linear decay. In the meta-testing, we use the same hyperparameters and train 10 epochs for finetuning.

All the compared models use the same templates (map the input to the text) and the same verbalizers (map the label to the text) from the Crossfit paper \cite{ye2021crossfit}, as they covered more classification datasets than other frameworks.  
Note that the choices of template/verbalizer could cause large variance in performance \cite{zhao2021calibrate}, and the effectiveness of Crossfit template/verbalizer had not been fully studied.  

We use two NVIDIA A5000 for our experiments. The supervised pretraining takes 3 days and the evaluation takes 1 week for all the compared baselines. 
\section{Details about Task Partition}
\label{sec:appendix}

\begin{table}[h]

\centering 
\resizebox{0.8\textwidth}{!}{
\normalsize     
\begin{tabular}{l|llll}
Datasets                 & Labels & Test sentences & Citation \\ \hline
glue-cola                     & 2      & 1043           &     \cite{cola}     \\
glue-qqp                      & 2      & 40430          &    \cite{qqp}      \\
glue-sst2                     & 2      & 872            &    \cite{sst2}      \\
glue-mrpc                     & 2      & 408            &    \cite{mrpc}      \\
scitail                       & 2      & 1304           &  \cite{scitail}        \\
amazon\_polarity              & 2      & 1000           &   \cite{amazon}       \\
ag\_news                      & 4      & 7600           &    \cite{zhang2015character}      \\
rotten\_tomatoes              & 2      & 1066           &   \cite{rotten}       \\
hate\_speech\_offensive       & 3      & 4957           &   \cite{hateoffensive}      
\end{tabular}}
\caption{The statistics of the 9 test data.}
\end{table}

\begin{lstlisting}[language=json,firstnumber=1,numbers=none]
{
    "Suprevised_pretraining": ["tweet_eval-stance_hillary", "ethos-sexual_orientation", "climate_fever", "hate_speech18", "tweet_eval-emotion", "hatexplain", "ethos-race", "emotion", "superglue-rte", "discovery", "anli", "wiki_auto", "scicite", "financial_phrasebank", "sms_spam", "kilt_fever", "tweet_eval-stance_climate", "medical_questions_pairs", "tweet_eval-stance_feminist", "ethos-directed_vs_generalized", "glue-wnli", "health_fact", "liar", "yahoo_answers_topics", "ethos-religion", "circa", "ethos-disability", "emo", "tweet_eval-hate", "tweet_eval-sentiment", "superglue-wic", "tweet_eval-emoji", "glue-qnli", "ade_corpus_v2-classification", "ethos-national_origin", "dbpedia_14", "poem_sentiment", "yelp_polarity", "tweet_eval-stance_atheism", "onestop_english", "glue-rte", "wiki_qa", "ethos-gender", "superglue-wsc", "tweet_eval-stance_abortion", "paws", "tweet_eval-offensive"],
    "meta_test": ["glue-cola", "glue-qqp", "glue-sst2", "glue-mrpc", "scitail", "amazon_polarity", "ag_news", "rotten_tomatoes", "hate_speech_offensive"]
}
\end{lstlisting}

\section{Additional results}

\begin{table*}[t]
\resizebox{0.95\textwidth}{!}{
\normalsize
\centering
\begin{tabular}{lcccccccccccc}
\toprule
 \textbf{Method} & \textbf{CoLA} & \textbf{QQP} & \textbf{SST-2} & \textbf{MRPC} & \textbf{SCITAIL}&  \textbf{Amazon} & \textbf{AGNews} & \textbf{rotten\_tomatoes} & \textbf{hate\_speech} & \textbf{AVG} \\ 
\midrule
$\text{Random-guess}$  & \mathnum{50.5} & \mathnum{49.8} & \mathnum{49.9} &  \mathnum{50.0} & \mathnum{49.8} & \mathnum{49.9} & \mathnum{24.0} & \mathnum{46.8} & \mathnum{34.1} & \mathnum{44.9} \\

\midrule
\multicolumn{2}{l}{\textit{PLM + $0$-shot}}\\
\midrule
$\text{Crossfit}$        &       \mathnum{0.0}   &       \mathnum{0.0}   &       \mathnum{0.0}   &       \mathnum{0.0}   &       \mathnum{0.0}   &       \mathnum{0.0}   &       \mathnum{0.0}   &       \mathnum{0.0}   &       \mathnum{0.0}   &\mathnum{0.0}  \\
$\text{Unifew}$  &       \mathnum{0.0}   &       \mathnum{0.0}   &       \mathnum{0.0}   &       \mathnum{0.0}   &       \mathnum{0.0}   &       \mathnum{0.0}   &       \mathnum{0.0}   &       \mathnum{0.0}   &       \mathnum{0.0}   &       \mathnum{0.0}  \\
$\text{EFL}$       &       \mathbnum{62.6}  &       \mathnum{48.1}  &       \mathnum{44.7}  &       \mathbnum{58.6}  &       \mathnum{44.8}  &       \mathnum{42.3}  &       \mathbnum{31.5}  &       \mathnum{45.4}  &       \mathnum{19.6}  &\mathnum{44.2}  \\
$\text{\ours}$ &       \mathnum{32.7}  &       \mathbnum{63.2}  &       \mathbnum{57.0}  &       \mathnum{31.6}  &       \mathbnum{50.4}  &       \mathbnum{55.9}  &       \mathnum{25.4}  &       \mathbnum{53.4}  &       \mathbnum{78.3}  &       \mathbnum{49.8}  \\
\midrule

\multicolumn{2}{l}{\textit{PLM + Supervised Pretraining + $0$-shot}} \\
\midrule
$\text{Crossfit}$        &       \mathnum{0.0}   &       \mathnum{0.0}   &       \mathnum{33.4}  &       \mathnum{0.0}   &       \mathnum{0.2}   &       \mathnum{9.9}   &       \mathnum{0.0}   &       \mathnum{59.9}  &       \mathnum{0.0}   &       \mathnum{11.5}  \\
$\text{Unifew}$  &       \mathnum{0.0}   &       \mathnum{0.0}   &       \mathnum{60.6}  &       \mathnum{0.0}   &       \mathnum{48.4}  &       \mathnum{63.7}  &       \mathnum{8.0}   &       \mathnum{57.4}  &       \mathnum{0.0}   &       \mathnum{26.5}  \\
$\text{EFL}$        & \mathnum{46.2}  &       \mathbnum{60.5}  &       \mathnum{79.1}   & \mathnum{33.1}  &       \mathnum{47.2}  &       \mathnum{71.9}  &       \mathnum{60.8}  &       \mathnum{72.5}  &       \mathnum{12.7}  &       \mathnum{53.8}  \\
$\text{\ours}$ &       \mathbnum{58.5}  &       \mathnum{45.3}  &       \mathbnum{83.0}  &       \mathbnum{58.1}  &       \mathbnum{68.7}  &       \mathbnum{89.7}  &       \mathbnum{52.8}  &       \mathbnum{78.1}  &       \mathbnum{34.3}  &       \mathbnum{63.2}  \\
\midrule
\multicolumn{2}{l}{\textit{PLM + $10$-shot fine-tuning}}\\

\midrule
$\text{Crossfit}$        &       \res{55.3}{5.0}         &       \res{53.4}{9.8}         &       \resb{81.2}{8.9}         &       \resb{60.0}{11.1}        &       \res{58.8}{5.4}         &       \res{87.9}{6.1}         &       \res{83.7}{6.6}         &       \res{65.0}{23.5}        &       \res{42.8}{14.4}        &   \mathnum{65.3}  \\
$\text{Unifew}$  &       \res{49.0}{4.9}         &       \resb{60.4}{6.0}         &       \res{71.2}{11.5}        &       \res{57.7}{6.3}         &       \res{53.4}{2.4}         &       \resb{88.8}{3.6}         &       \resb{86.5}{1.8}         &       \resb{73.4}{9.5}         &       \res{34.9}{6.8}         &       \mathbnum{63.9}  \\
$\text{EFL}$       &       \resb{63.7}{0.2}         &       \resb{60.4}{0.2}         &       \res{79.5}{0.2}         &       \res{32.3}{0.4}         &       \res{46.7}{1.1}         &       \res{72.0}{0.0}         &       \res{62.3}{0.6}         &       \res{72.4}{0.2}         &       \res{13.8}{0.6}         &   \mathnum{55.9}  \\

$\text{\ours}$ &       \res{38.6}{4.4}         &       \res{55.6}{3.5}         &       \res{62.4}{3.4}         &       \res{46.1}{2.4}         &       \resb{59.1}{2.7}         &       \res{64.0}{1.6}         &       \res{57.4}{2.3}         &       \res{61.8}{2.0}         &       \resb{58.6}{11.0}        &       \mathnum{55.9}  \\
\midrule

\multicolumn{2}{l}{\textit{PLM + Supervised Pretraining + $10$-shot fine-tuning}}\\
\midrule
$\text{Crossfit}$         &       \res{25.7}{25.1}       &       \res{13.6}{18.2}       &       \res{80.6}{4.0}        &       \res{28.1}{28.7}       &       \res{53.9}{11.7}       &       \res{85.2}{6.5}      &       \res{31.7}{17.7}       &       \res{75.8}{1.2}        &       \res{2.4}{3.2}         &       \mathnum{44.1}  \\
$\text{Unifew}$  &       \res{46.6}{12.4}       &       \res{37.5}{30.6}       &       \res{60.9}{1.4}        &       \res{23.8}{24.9}       &       \res{51.0}{1.8}        &       \res{63.9}{2.9}      &       \res{52.2}{16.1}       &       \res{57.9}{1.5}        &       \res{9.8}{6.5}         &       \mathnum{44.8}  \\
$\text{EFL}$  &     \res{46.1}{0.0}        &       \resb{60.4}{0.0}        &       \res{79.1}{0.1}        &       \res{33.1}{0.0}        &       \res{47.2}{0.1}        &       \res{72.0}{0.0}        &       \res{60.9}{0.0}        &       \res{72.5}{0.0}       &\res{12.9}{0.1}        &       \mathnum{53.8}  \\
$\text{\ours}$  &       \resb{60.5}{0.6}        &       \res{51.8}{1.9}        &       \resb{83.2}{0.2}        &       \resb{69.9}{0.9}        &       \resb{71.0}{0.9}        &       \resb{89.4}{0.1}      &       \resb{70.3}{2.1}        &       \resb{78.7}{0.2}        &       \resb{44.7}{2.2}        &       \mathbnum{68.8}  \\
\bottomrule
\end{tabular}
}
\caption{The complete table of the main result. 
}
\label{tab:backup}
\end{table*}



\end{document}